# Praaline: Integrating Tools for Speech Corpus Research

## George Christodoulides

Centre Valibel, Institute for Language & Communication, University of Louvain
Place Blaise Pascal 1, B-1348 Louvain-la-Neuve, Belgium
E-mail: george@mycontent.gr

### Abstract

This paper presents Praaline, an open-source software system for managing, annotating, analysing and visualising speech corpora. Researchers working with speech corpora are often faced with multiple tools and formats, and they need to work with ever-increasing amounts of data in a collaborative way. Praaline integrates and extends existing time-proven tools for spoken corpora analysis (Praat, Sonic Visualiser and a bridge to the R statistical package) in a modular system, facilitating automation and reuse. Users are exposed to an integrated, user-friendly interface from which to access multiple tools. Corpus metadata and annotations may be stored in a database, locally or remotely, and users can define the metadata and annotation structure. Users may run a customisable cascade of analysis steps, based on plug-ins and scripts, and update the database with the results. The corpus database may be queried, to produce aggregated data-sets. Praaline is extensible using Python or C++ plug-ins, while Praat and R scripts may be executed against the corpus data. A series of visualisations, editors and plug-ins are provided. Praaline is free software, released under the GPL license.

## 1. Introduction

This paper presents *Praaline*, an open-source software system for managing, annotating, analysing and visualising speech corpora. It attempts to address the needs of researchers working with speech corpora, who are often faced with multiple tools and formats and need to work with ever-increasing amounts of data in a collaborative way.

Praaline is based on and extends *Praat* (Boersma & Weenink 2014) and *Sonic Visualiser* (Cannam et al. 2010), and interfaces with the *R* statistical language (R Core Team 2013). Instead of creating a system from scratch, we have chosen to focus on the integration of open-source tools that are already widely-used in the community. As a result, researchers using *Praaline* can benefit from the features, extensions and contributions to these tools. This design allows for the reuse of many existing tools and scripts providing automated annotation and analyses.

Praaline is written in C++ using the Qt framework (Nokia) for both its core functions and user interface. It is cross-platform software that runs under Windows, Linux and Mac.

Recordings and corpus annotations may reside in a file system (cf. infra) and are managed through a relational database. It is possible to use SQLite for local installations or MySQL for client-server access. *Praaline* can import and export annotations in different formats, including *Praat* TextGrids, *TranscriberAG* (Barras et al., 1998), *ELAN* (Brugman & Russel, 2004) and *EXMARaLDA Partitur* (Schmidt, 2012) files. An integrated user interface permits the management of corpus data and metadata, annotation using editors or automated procedures, visualisation for the purposes of data exploration or demonstration, and querying the data for analysis. Praaline can be extended with plug-ins written in C++ or Python, and also supports executing scripts written in the *Praat* scripting language or for the *R* system against the corpus data and annotations.

## 2. Corpus Management

Users may construct their corpora from scratch or by importing a set of existing files (e.g. recordings and annotations). The user interface for defining and managing a corpus is shown in Figure 1. A corpus is organised in Communications (communicative situations) and Speakers. Each Communication may consist of several Recordings and Annotations. Speakers participate in Communications with specific Roles; and Annotations contain Annotation Levels (see section 3).

Using these basic concepts, users define the Annotation Structure and the Metadata Structure of their corpora. These definitions specify the information stored for each type of object. Corpus definitions are stored in a relational database and/or in XML files. It is possible to import and export the XML formats of *EXMARaLDA's* Coma editor. Users may also create projects and collaborate with colleagues using version control (i.e. compare and merge changes made to the corpus metadata and annotation).

Corpus metadata can be used to define sub-corpora, and selectively apply automated annotation procedures or queries to them, in other parts of the system. Aggregate data (e.g. number of samples, total recording time, counts of specific units like tokens or syllables) may be exported.

## 3. Annotation

While following the classical timeline model for annotating speech data, Praaline offers a key improvement: the option to define structural links between the annotation tiers, such as hierarchy, containment, attachment, controlled vocabularies etc. The annotation includes knowledge about the relationships between the different layers. An Annotation Level may contain any number of attributes, and relationships between Annotation Levels are encoded as part of the corpus structure. Praaline does not impose any specific set of Annotation Levels or attributes (sample sets are included to help users). Annotation attributes have associated data types, and may be marked as optional.

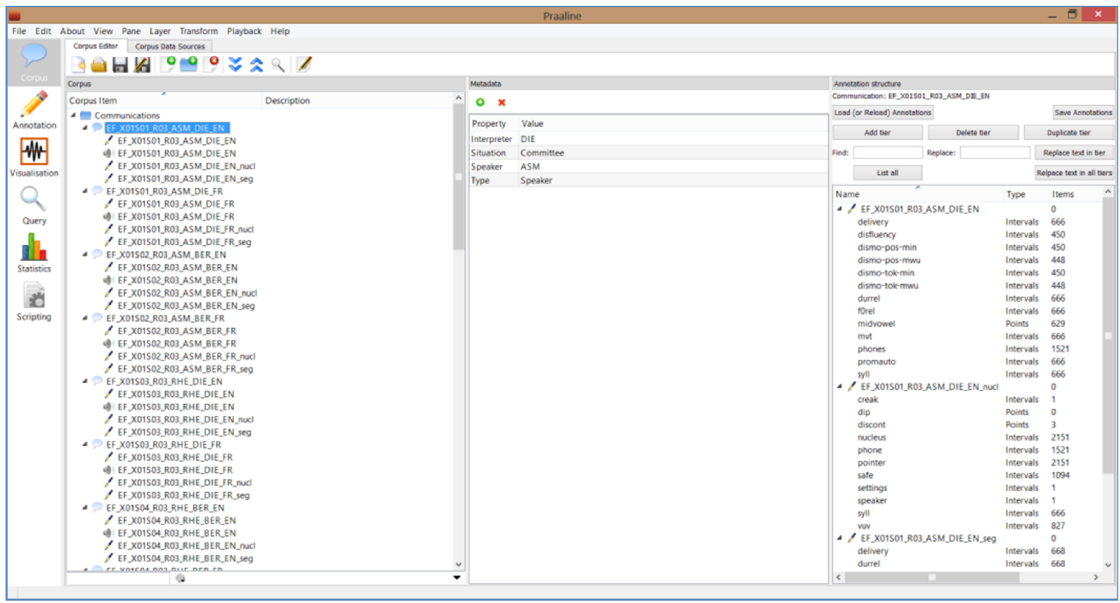

Figure 1: Corpus management mode (left pane: corpus structure; middle: metadata; right: annotation structure)

Annotations may be imported, entered manually using an editor, or obtained automatically by applying annotation plug-ins. *Praaline* allows the user to apply a cascade of annotation plug-ins on the entire corpus, or on subsets of it. Heterogeneous annotation utilities can be applied sequentially on the corpus. For example, a compiled plug-in for feature extraction, may be followed by a *Praat* script for prosodic annotation, and then by a POS tagger and an NLP parser in Python, while finally an R script is used to perform a statistical analysis. *Praaline* handles the data conversions needed to allow such combinations.

The Annotation Level and Annotation Attribute can represent timeline annotations, and also ensure integrity of the data. It is often the case that congruent annotation tiers are used (e.g. in *Praat*) to represent multiple features of the same object (e.g. a syllable, and an indication of whether it was perceived as prominent or not, or whether it is disfluent). While practical for small amounts of data, this system quickly leads to problems when corpora get larger: e.g. discrepancies in tier boundaries that should had been aligned according to the model; or data incoherence between tiers that are supposedly linked (e.g. phones-syllables). Since these relationships are explicitly captured explicitly in *Praaline*, it is possible to check the data integrity of a set of corpus annotations, possibly correcting them automatically.

A spreadsheet-like editor allows the user to simultaneously edit attributes belonging to several different Annotation Levels. For example, in a corpus for prosodic studies, we may discern at least three levels: phones, syllables and tokens (words). Each level may have a number of associated Attributes (e.g. syllables may be described by several automatically extracted prosodic features). The editor (Figure 2) allows the user to view and update a selection of attributes from each level; it is essentially a vertical timeline display, synchronised with the sound signal. It is also possible to define bookmarks in the corpus, and move directly to these points.

Figure 2: Vertical timeline editor

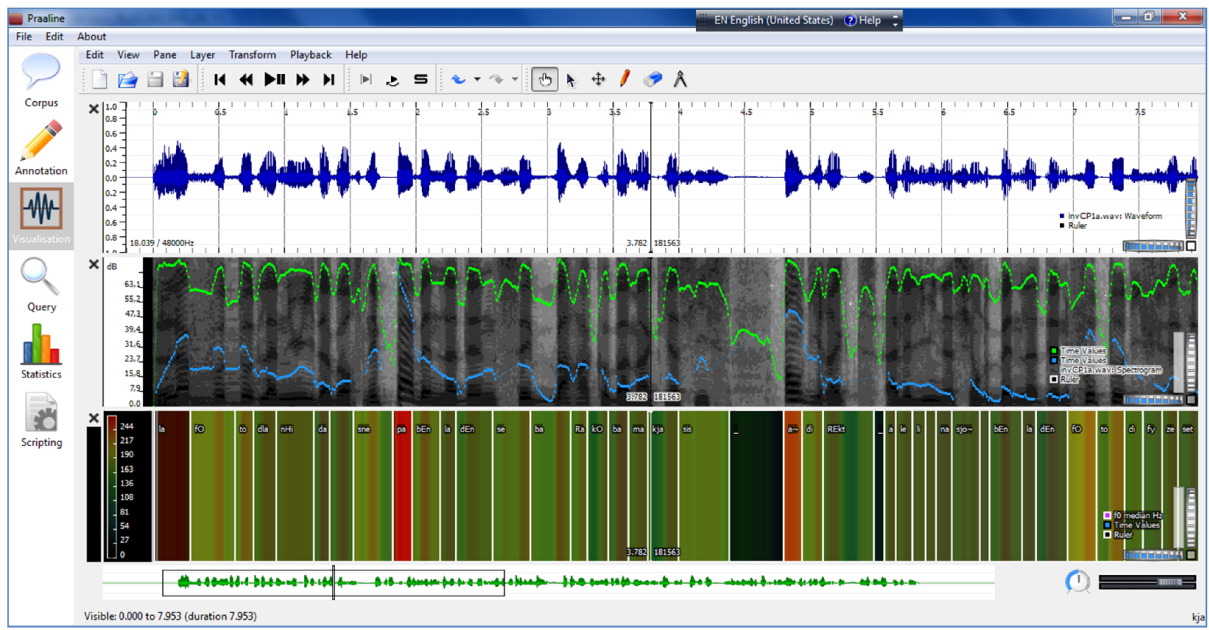

Figure 3: A visualization of prominent syllables with colour coding

The Annotation Level – Annotation Attribute data model translates directly into a relational SQL database. Each Annotation Level is a table, and Annotation Attributes are columns. *Praaline* uses this system to provide querying functionality (see section 5, infra). In addition to the possibilities offered by the user interface and through scripting, an advanced user can directly query *Praaline's* SQL database. It is important to note that the schema is dynamic and adapted to each corpus definition (with the exception of system tables that are always present). Finally, it is envisaged that this corpus metadata and annotation database can be linked to web interface to provide outside users with limited access to the corpus.

## 4. Visualisation

The visualisation module of Praaline is based on *Sonic Visualiser* (Cannam et al., 2010). Visualisations can display waveforms, spectrograms, melodic spectrograms, any combination of annotation levels and tiers, numerical data (points, curves, histograms, colour-coded regions etc.). Plug-ins may add visualisations: for example, we have adapted *Prosogram* (Mertens, 2004) do display prosodic analysis information.

These elements can be combined to present annotations in a format appropriate for each type of investigation. For example, a dialogue involving multiple speakers can be visualised in speaker turns. In Figure 3, we show a colour-coded annotation of prominent syllables (along with a waveform, spectrogram, intensity, f0, and transcription). As can be seen in Figure 4, in order to study the prosody of simultaneous interpreting, we used a *Praaline* visualisation comparing the prosodic characteristic of the speaker's and the interpreter's speech, based on Mertens (2004), and calculating similarity measures (cf. DeLooze & Rauzy 2011; Christodoulides 2013).

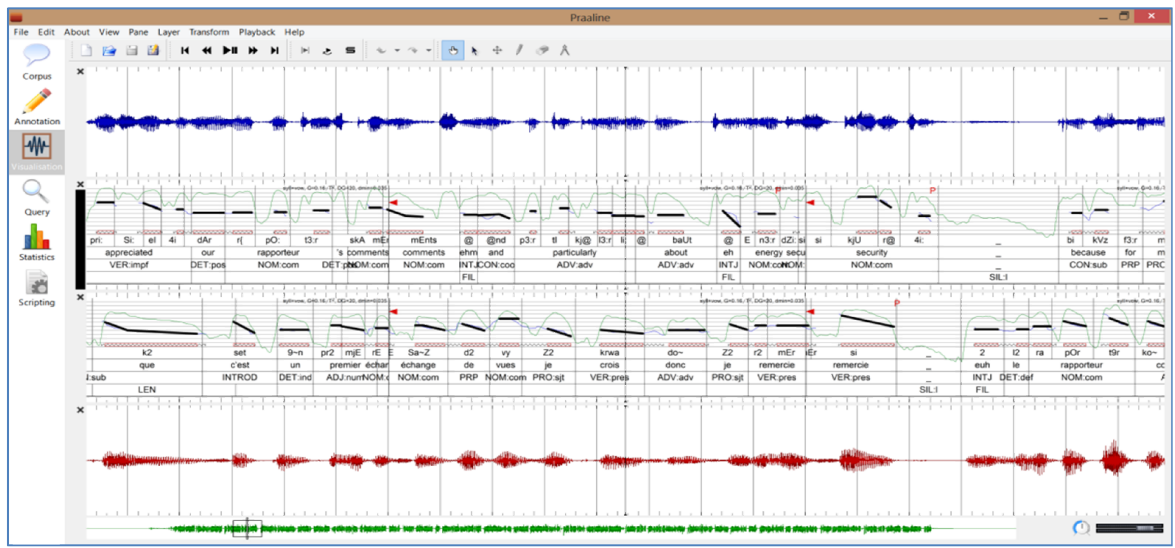

Figure 4: A visualisation for comparing the prosody of a speaker and an interpreter using
a multi-channel recording and Prosogram (Mertens 2004)

## 5. Querying

Corpus annotations are stored in a relational database, the schema of which is dynamically constructed based on the annotation structure definition. Annotation levels correspond to database tables and annotation tiers to columns. The relationships between the different levels are also encoded. Praaline simplifies the conversion of structured annotation into two-dimensional tables suitable for statistical analysis. SQL queries can be used to select and summarise a subset of the corpus.

It is possible to construct a Dataset by an interactive query editor. For each attribute, functions can be applied to calculate aggregate measures (e.g. sum, mean, standard deviation, etc.); a filter can be used to limit the returned values; and a normalisation transformation (e.g. z-score over all samples of the same sub-corpus) may be applied. In this way, researchers may more easily explore and analyse the information in the corpus, in an interactive way and without necessarily resorting to scripts.

Furthermore, concordances can be extracted using Praaline, based on simple value filters or regular expressions (search term, left and right context). The results of such queries are objects that can be further processed, using the statistical analysis module, or exported for use with other software.

## 6. Statistical Analysis

Praaline interfaces with the *R* statistical environment, through the *Rcpp* package (Eddelbuettel & François 2011). Corpus annotations (as well as the results of corpus queries) are exposed to R as data frames, allowing for the use of R commands, scripts and extensions to analyse the data. *Praaline* provides a two-way link between the corpus and *R*: the results of analyses performed using *R* can be posted back into the corpus database, by adding to or updating an existing annotation level, or by creating a new annotation level.

## 7. Programmability

*Praaline* can be extended with plug-ins, written in C++ using a simple API and compiled. This method is suitable for plug-ins adding substantial functionality to the system. Praaline is also scriptable with Python, by providing bindings to its core functionality. *Praat* scripts can be executed, in which case the corpus data are available as (virtual) Praat objects. Finally it is possible to evaluate *R* scripts accessing corpus data in the form of data frames. A programmer may mix the following techniques in order to use the tool that is best suited to each task. Currently available plug-ins include the following:

- An adapted version of Mertens' (2004) Prosogram.
- An automatic rule-based syllabifier, based on the increasing sonority principle (a list of allowed syllable onsets for each language is required).
- A plug-in version of the DisMo morphosyntactic annotator, for English and French (Christodoulides et al., 2014).
- A plug-in for calculating similarity and convergence measures in dialogue, based on the methodology of De Looze & Rauzy (2011).

## 8. Conclusion and Future Work

Praaline is currently under active development, and is made available to the research community under the GPL licence. It can be downloaded from **www.praaline.org**.

We welcome feedback on the functionality and future directions of the project. We are exploring the development of a bridge between *Praaline* databases and open-source content management systems, in order to facilitate the publishing of corpus data to the web.

## 9. References

Barras, C.; Geoffrois, E.; Wu, Z. and Liberman, M. (1998). Transcriber: a Free Tool for Segmenting, Labeling and Transcribing Speech. In Proceedings of LREC 1998, pp. 1373-1376.

Boersma, P.; Weenink, D. (2009). Praat: doing phonetics by computer. http://www.praat.org

Brugman, H. and Russel, A. (2004). Annotating Multimedia/ Multi-modal resources with ELAN. In *Proceedings of LREC 2004*.

Cannam, C.; Landone, C.; Sandler M. (2010). Sonic Visualiser: An open source application for viewing, analysing, and annotating music audio files, Proceedings of the ACM Multimedia 2010 International Conference, pp. 1467-1468.

Christodoulides, G. (2013): Prosodic features of simultaneous interpreting. In Mertens, P. & A.C. Simon (Eds), *Proceedings of the Prosody-Discourse Interface Conference 2013 (IDP-2013)*, pp. 33-37.

Christodoulides, G.; Avanzi, M. and Goldman, J-Ph. (2014). DisMo DisMo: A Morphosyntactic, Disfluency and Multi-Word Unit Annotator. An Evaluation on a Corpus of French Spontaneous and Read Speech, In *Proceedings of LREC 2014*.

De Looze C.; Rauzy S. (2011). Measuring speakers' similarity in speech by means of prosodic cues: methods and potential. In *Proceedings of Interspeech 2011*, pp. 1393-1396.

Eddelbuettel, D; François, R. (2011). Rcpp: Seamless R and C++ Integration. Journal of Statistical Software, 40(8), pp. 1-18.

Mertens, P. (2004). The Prosogram: Semi-Automatic Transcription of Prosody based on a Tonal Perception Model. In B. Bel & I. Marlien (Eds.) *Proceedings of Speech Prosody 2004*, Nara, Japan.

R Core Team (2013). R: A language and environment for statistical computing. R Foundation for Statistical Computing, Vienna, Austria. http://www.R-project.org.

Schmidt, T.; Wörner, K. (2009). EXMARaLDA – Creating, analysing and sharing spoken language corpora for pragmatic research. Pragmatics 19(4), pp. 565–582.